\documentclass[10pt,journal,compsoc]{IEEEtran}

\ifCLASSOPTIONcompsoc
  \usepackage[nocompress]{cite}
\else
  \usepackage{cite}
\fi

\usepackage{pgfplots}
\usepackage{float}
\usepackage{subfig}
\usepackage{graphicx}
\usepackage{tabularx,ragged2e,booktabs}

\usepackage{algorithm}
\usepackage{algpseudocode}
\usepackage{amsmath}
\usepackage{graphics}
\usepackage{epsfig}
\usepackage{soul}

\definecolor{SRCF}{HTML}{e68d61}
\definecolor{EARLIEST}{HTML}{5cc2b7}
\definecolor{TEASER}{HTML}{67abd6}
\definecolor{ETEeTSC}{HTML}{c3b0d6}
\definecolor{SPN}{HTML}{e76781}
\definecolor{MDDNN}{HTML}{89d78c}

\ifCLASSINFOpdf
 
\else
 
\fi

\begin{document}

\title{Snippet Policy Network for Multi-class Varied-length ECG Early Classification}

\author{Yu Huang,
        Gary G. Yen,~\IEEEmembership{Fellow,~IEEE,}
        Vincent S. Tseng,~\IEEEmembership{Fellow,~IEEE} 
\IEEEcompsocitemizethanks{
\IEEEcompsocthanksitem Yu Huang is with the Department of Computer Science, National Yang Ming Chiao Tung University, Hsinchu, R.O.C., e-mail: yuvisu.cs04g@nctu.edu.tw
\IEEEcompsocthanksitem Gary G. Yen is with the School of Electrical and Computer Engineering, Oklahoma State University, Stillwater, OKlahoma, USA., e-mail: gyen@okstate.edu
\IEEEcompsocthanksitem Vincent S. Tseng is with the Department of Computer Science, National Yang Ming Chiao Tung University, Hsinchu, R.O.C., e-mail: vtseng@cs.nctu.edu.tw}
\thanks{Corresponding author: Vincent S. Tseng}
}

\markboth{
}%
{Huang \MakeLowercase{\textit{et al.}}: Snippet Policy Network for Multi-classVaried-length ECG Early Classification}

\IEEEtitleabstractindextext{%
\begin{abstract}

Arrhythmia detection from ECG is an important research subject in the prevention and diagnosis of cardiovascular diseases. The prevailing studies formulate arrhythmia detection from ECG as a time series classification problem. Meanwhile, early detection of arrhythmia presents a real-world demand for early prevention and diagnosis. In this paper, we address a problem of cardiovascular diseases early classification, which is a varied-length and long-length time series early classification problem as well. For solving this problem, we propose a deep reinforcement learning-based framework, namely Snippet Policy Network (SPN), consisting of four modules, snippet generator, backbone network, controlling agent, and discriminator. Comparing to the existing approaches, the proposed framework features flexible input length, solves the dual-optimization solution of the earliness and accuracy goals. Experimental results demonstrate that SPN achieves an excellent performance of over 80\% in terms of accuracy. Compared to the state-of-the-art methods, at least 7\% improvement on different metrics, including the precision, recall, F1-score, and harmonic mean, is delivered by the proposed SPN. To the best of our knowledge, this is the first work focusing on solving the cardiovascular early classification problem based on varied-length ECG data. Based on these excellent features from SPN, it offers a good exemplification for addressing all kinds of varied-length time series early classification problems.

\end{abstract}

}

\maketitle

\IEEEdisplaynontitleabstractindextext

\IEEEpeerreviewmaketitle

\IEEEraisesectionheading{\section{Introduction}\label{sec:introduction}}

\IEEEPARstart{I}{n} recent years, the incidence of cardiovascular diseases (CVDs) has practically exploded, which has become a significant threat to human life due to high mortality. Continuous monitoring of CVDs for patients well in advance has been proven an effective measure to save lives. Electrocardiogram (ECG) \mbox{\cite{van2004clinical}} is a common non-invasive measurement that reflects the physiological state of the heart, and it is one of the most important diagnostic tools in the current age. With the development of smart wearable devices in recent years, patients can acquire ECG devices ubiquitously for personal healthcare monitoring. Although ECG signals are convenient to collect, it remains challenging for medical professionals and cardiologists to analyze such multifarious data. Hence, automatic ECG classification modeling has become an important topic in the research community.

Time series is a common type of data format for representing ECG signals, a collection of values presenting signal strengths ordered by timestamp sequentially. ECG classification or cardiovascular diseases detection problem can be formulated as a time series classification problem. It is a problem of assigning one of a predefined class to a time series, e.g., classifying a signal of ECG motions as normal or an impending atrial fibrillation \cite{yao2020multi} or classifying signals from different patients with or without chronic obstructive pulmonary diseases \cite{alkukhun2014electrocardiographic} \cite{GOUDIS2015264}. Conventional time series classification works on a given fixed-length time series and assumes accessing the entire input time series at making a decision. Unfortunately, it cannot meet real-world requirements in many scenarios. For example, in the ICU because different patients are monitored at different times and duration, this leads to generating varied-length time series data. In time-sensitive applications, making a decision as early as possible is crucial for improving practicability. For instance, early diagnosis can provide patients with timely and effective treatment, which is vital as many heart diseases are fatal in a short time. As a result, early time series classification for ECG signals is an important research issue. In this study, we focus on solving the early time series classification problem, which aims to early classify a time series with confidence by seeing as few data as possible.

\begin{figure}
    \centering
    \includegraphics[width=\linewidth]{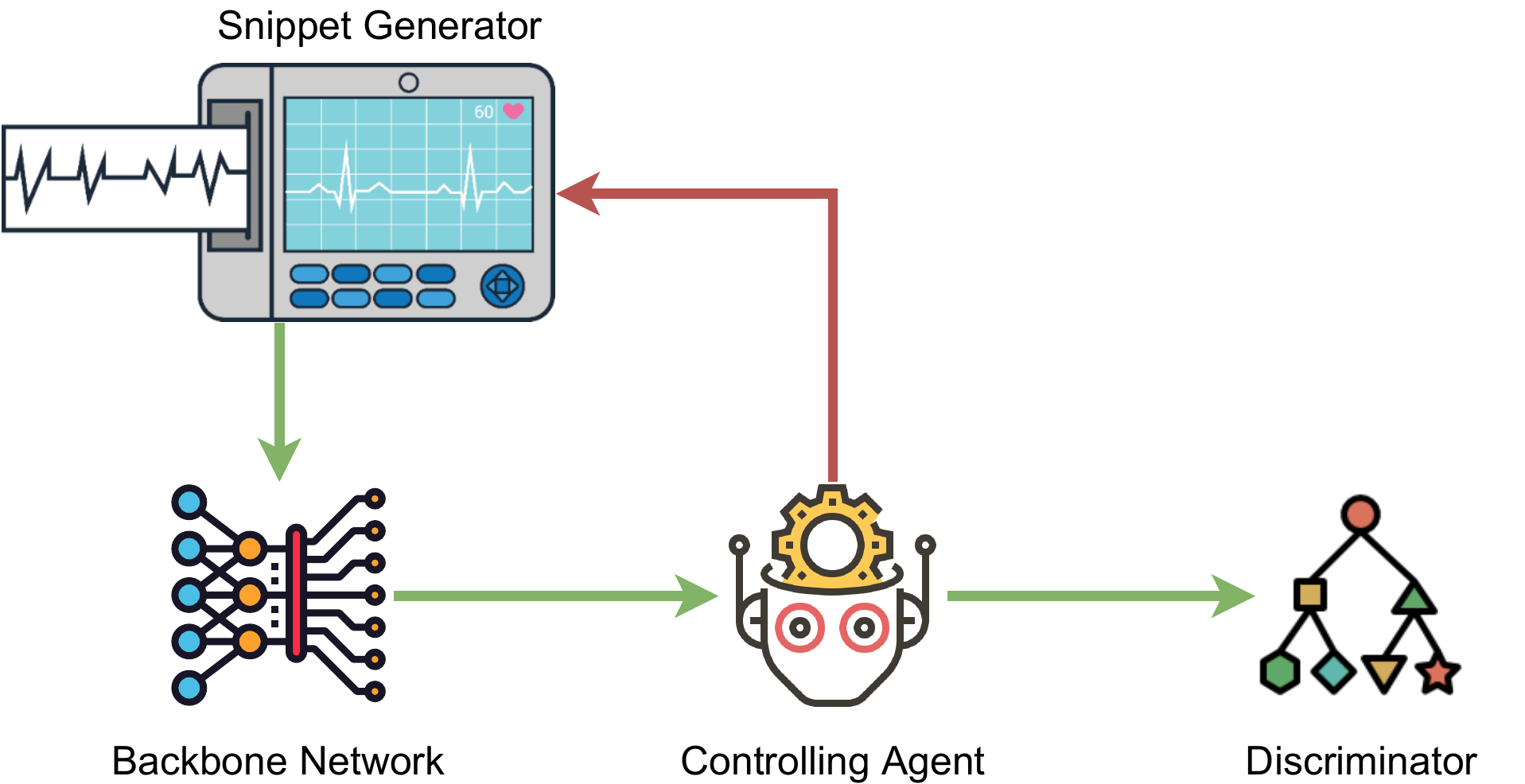}
    \caption{The  proposed snippet policy network contains four modules, which are snippet generator, backbone network, controlling agent and discriminator.}
    \label{fig:SPN}
\end{figure}

Early time series classification (eTSC) has two distinct goals: classifying a given time series early and accurately. However, the earlier the time series has to be classified, the fewer data points are provided for this task, which usually leads to lower accuracy. In contrast, the higher the classification accuracy is desired, the more measurements are needed, and the decision would be made at a later time. Apparently, these two goals are contradictory in nature. It is a highly challenging issue of early time series classification in determining the earliest point when an incoming time series can be correctly classified. Take patient care as an example; collecting more physical signals of the patients to be analyzed naturally leads to a higher accurate diagnosis, but may miss valuable time preparing an emergency plan for saving a life.

Most existing approaches in eTSC, such as \cite{wang2016earliness, mori2017reliable, xing2012early, xing2011extracting}, assume that all time series being classified have a defined start time. Consequently, these methods assume that characteristic patterns appear roughly at the same offset in all time series, and try to find the fixed fraction of each time series in the training pool. In such settings, the threshold of earliness depends on the average accuracy of classification, which implies that it is not designed to find the optimal solution. For example, monitoring sensors start their observations at arbitrary points in time of a time series for different patients, and a characteristic pattern would appear at arbitrary points in time of a time series. Intuitively, existing methods are expected to analyze a time series when the observed devices start simultaneously and the characteristic patterns appear simultaneously; while in CVDs monitoring applications, the ECG signals would be collected at an arbitrary time. In this case, it is sub-optimal to use the existing methods for solving ECG early classification problem. 

In this paper, to address the above issues, we propose a snippet policy network (SPN), as shown in Figure \ref{fig:SPN}, which embodies a deep reinforcement neural network model  to learn multivariate information from varied-length ECG signals. The proposed framework consists of four modules: snippet generator, backbone network, controlling agent, and discriminator. To well model the ECG signals regarding the periodicity, we use the snippet generator to split the original time series into snippet series. Then, to capture inner-snippet spatial dependency and inter-snippet temporal dependency, the backbone network that combines convolutional and recurrent networks is built to learn hidden representations. To find an appropriate prediction time point, an agent controls the backbone network's whole process and informs the discriminator. Finally, the discriminator receives the controlling agent's notification and produces the predictive result according to the hidden representations from the backbone network.

The contributions of this paper are the following:
\begin{itemize}
    \item We address the problem of classifying cardiovascular diseases based on varied-length ECG signals as early as possible, which has not been well explored in the research community.
    \item To solve the problem of ECG early classification, we propose a novel deep reinforcement learning framework consisting of four modules, snippet generator, backbone network, controlling agent, and discriminator.
    \item The controlling agent solves the bi-objective optimization problem solution of the earliness and accuracy goals in conflict. It allows the model to classify CVDs accurately as well as to find the earlier prediction point in time.
    \item The proposed model is evaluated on a public CVDs classification dataset and has demonstrated its performance in classifying different cardiovascular diseases. Experimental results show that the proposed framework outperforms existing state-of-the-art approaches.
\end{itemize}

The rest of this paper is organized as follows: Section 2 briefly outlines the existing related works. Section 3 introduces the architecture of the proposed SPN framework, while the experimental evaluation results are described in Section 4. The conclusion of this paper is summarized in Section 5.

\section{Related Work}
The early time series classification problem has attracted researchers from data mining and machine learning communities in the past decade. According to the strategy of the existing eTSC approaches used, they can be broadly divided into two different branches, which are feature-based methods and series-based methods. Feature-based methods extract meaningful patterns and exploit these patterns to build the early classifier. On the other hand, series-based methods use the raw time series directly to learn a classification model.

\subsection{Feature-based Methods}
In 2011, Xing et al.\cite{xing2011extracting} developed the method, called Early Distinctive Shapelet Classification (EDSC), for extracting meaningful patterns in the eTSC problem. They proposed an interpretable feature called local shapelets, which are essentially sub-sequences of time series. Sub-sequences of time series are intuitive to present the physical meaning to end-users, and can effectively capture the local similarity among time series, so they have high interpretability. Ghalwash et al.\cite{ghalwash2012early} generalized the definition of local shapelets to a multivariate context and proposed a method for early classification of multivariate time series accordingly. The proposed method, Multivariate Shapelets Detection (MSD), extracts patterns from all dimensions of the time series, which is called multivariate shaplet. A multivariate shapelet consists of multiple segments, where each segment is extracted from exactly one dimension. To extract interpretable patterns from multivariate time series data, Ghalwash et al.\mbox{\cite{ghalwash2013extraction}} proposed an optimization-based method for building predictive models on multivariate time series and mining relevant temporal interpretable patterns for early classification (IPED). The IPED method extracts a full-dimensional shapelet for each class from the binary matrix by solving a convex-concave optimization problem. The imbalanced class dataset in classification is a classical problem in the data mining field, and it remains a challenge on early time series classification problems. In 2019, He et al.\cite{he2019ensemble} proposed an adaptive ensemble framework to learn an early classification model on imbalanced multivariate time series data. The proposed ensemble framework was designed based on combining the multiple under-sampling approaches, dynamical subspace generation method, cluster-based shapelet selection method, and associate-pattern mining approach to deal with the implicit issues of inter-class and intra-class imbalances. These methods build an eTSC model by extracting features, resulting in increased computing complexity. Moreover, they achieve poor performance when facing long time series data. Our proposed method adopts the snippets concept and adaptive neural architecture to solve these issues mentioned above, which can effectively and efficiently handle long-time series early classification problems.

\subsection{Series-based Methods}

Xing et al.\cite{xing2012early} published the first work to introduce the problem of eTSC from a series-based approach. The authors developed the ECTS model based on the 1-nearest neighbor (1-NN) approach and the concept of minimum prediction length (MPL). In this paper, time series with the same 1-NN are clustered at first. The optimal prefix length for each cluster is calculated by analyzing the stability of the 1-NN decision for increasing data points in time. Then, the 1-NN approach is adopted to search among the clusters and label the class for each time series. In 2013, Parrish et al.\cite{parrish2013classifying} presented a method, called RelClass, based on quadratic discriminant analysis (QDA). A reliability score is defined as the probability that the predicted class for the truncated and the whole time series will be the same. At each timestamp of a given time series, RelClass checks if the reliability is higher than a user-defined threshold. Mori et al.\cite{mori2017early} proposed an early classification framework, SR2-CF2, based on combining a set of probabilistic classifiers and a stopping rule, designed by minimizing the earliness cost and accuracy cost. The method is conceptually simple and does not require complex parameter settings. These pioneer works started the early time series classification trend, but they left some research issues to be further addressed, such as multivariate time series and varied-length time series early classification problems. Mori et al. \cite{mori2017reliable} trained classifiers at specific timestamps, i.e., at percentages of the full-time series length. It learns a safe timestamp as the fraction of the time series, which states the model's best prediction timing. Furthermore, a reliability threshold is learned using the difference between the two highest class probabilities. Only predictions passing this threshold after the safe timestamp are chosen. Schafer and Leser \mbox{\cite{schafer2019teaser}} pointed out that time series is the non-fixed length in many real-world scenarios. To address this challenge, they proposed an ensemble framework, called Two-tier Early and Accurate Series classifiER (TEASER), for solving eTSC. TEASER’s decision for a prediction is treated as a classification problem, in which master classifiers continuously analyze the output of probabilistic, while slave classifiers decide if their results should be trusted or not.

In 2016, Wang et al.\cite{wang2016earliness} proposed the first deep learning-based early time series classification framework that leverages the information at different scales and captures the interpretable features at a very early stage. For handling the multivariate time series early classification problem, Huang et al.\cite{huang2018multivariate} devised a deep learning framework based on combining convolutional neural networks and long short-term memory with learning feature representation and relationship embedding in the extended sequences with time lags. Additionally, Martinez et al.\cite{martinez2018deep} addressed the early time series classification task with a novel approach based on reinforcement learning. The authors introduced an early classifier agent, an end-to-end reinforcement learning that can perform early classification efficiently. In order to improve the interpretability of the deep learning-based model for eTSC, Hsu et al.\mbox{\cite{hsu2019multivariate}} adopted an attention mechanism to identify the critical segments related to model performance, providing a base to facilitate a better understanding of the model. The above works, including our published related works, solve the early classification problem using fixed-length time series or sub-time series. Under real-world scenarios, the length of ECG signals varies length, and they are collected at different points in time. In this circumstance, the existing approaches cannot meet the real-world mission of early time series classification. In 2019, Hartvigsen et al.\cite{10.1145/3292500.3330974} proposed a reinforcement learning framework, called EARLIEST, for solving early classification problems and providing early halting point function. Russwurm et al.\cite{russwurm2019end} proposed a generic, end-to-end trainable framework for early time series classification (ETEeTSC). This framework embeds a learnable decision mechanism used in some existing models, such as Convolutional Neural Network (CNN) and Recurrent Neural Network (RNN). Many emerging approaches, including deep learning and reinforcement learning-based models, have been proposed to tackle different eTSC problems, but they still fall short when applying to real-world applications, e.g., CVDs early classification problems. As some critical issues remain to be satisfactorily addressed, such as long time series and varied-length time series, we propose in this paper a novel framework based on deep learning architecture for solving eTSC with the issues we mentioned above.

\begin{figure*}[h]
    \centering
    \includegraphics[width = \linewidth]{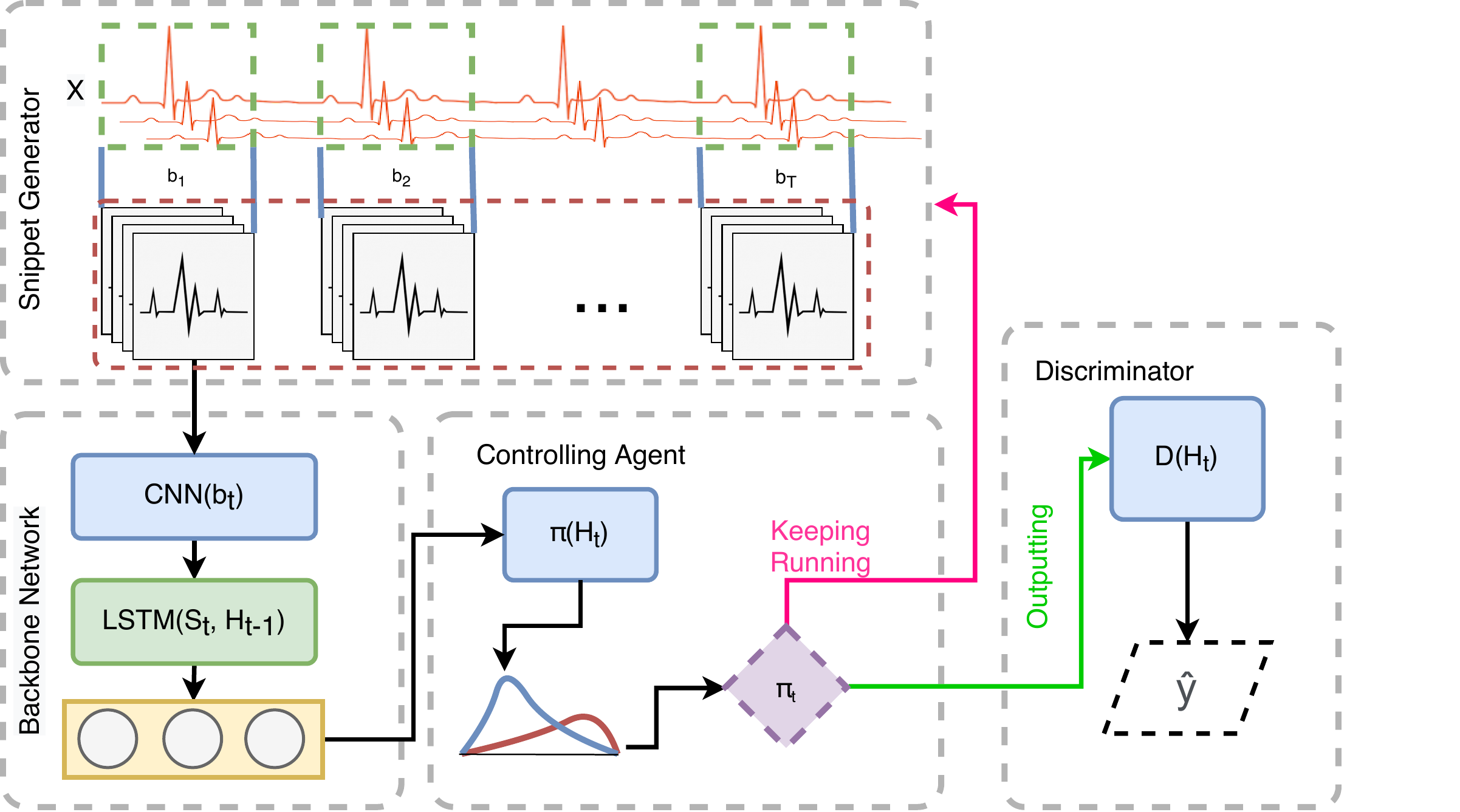}
    \caption{Overview of the proposed SPN. The input time series $X$ is split into a snippet, by the snippet generator. A selected snippet $b_{t}$ is fed to the backbone network sequentially, generating the hidden states $H_{t}$. Hidden state $H_{t}$ is then passed to the controlling agent to decide whether to keep running or to output the result, according to the intermediate parameterized distribution $\pi_{t}$ . The controlling agent informs the snippet generator feeding the next snippet into the backbone network if keeping running. Otherwise, the hidden state is passed into the discriminator and then producing the final result $\hat{y}$.}
    \label{fig:framework}
\end{figure*}

\section{Methodology}
\subsection{Problem Formulation}
ECG early classification is a real-world application of the multi-varied early time series classification problem. A multi-varied time series, such as a multi-lead ECG signal, reflects a specific individual's physiological status.  In this problem, a model is mandated to predict the correct class from varied-length ECG records as early as possible. Given a set of labeled multivariate time series (ECG signals), $D = (X, y)$ containing $N$ time series instances, $X^{(i)},i=1,…,N$ and the corresponding labels, $y^{(i)},i=1,…,N$. For each sample $X^{(i)}=\{x_1^i,x_2^i,…,x_L^i\}$, where $x_l^i$ contains $M$ variables recorded at timestamp $l$ and $L$ indicates the length of a given time series. The aim of eTSC is to learn a model, $f_\theta$, with parameter $\theta$, which generates a label $\hat{y}$ from time series $X$ with $l < L$. 

\subsection{Model Architecture}
The proposed model, named Snippet Policy Network (SPN), is constructed with three functional parts. First, to generate snippets of given ECG signals, second, to model ECG snippets for classification, and third, to select a halting snippet to make a task-dependently appropriate prediction. In detail, the snippet policy network is a deep neural network consisting of four sub-modules: (1) Snippet generator to split multivariate time series as sub-series, called snippets; (2) Backbone network to model multivariate snippets based on the output of snippet generator continuously, generating hidden representations; (3) Controlling Agent controls the whole process at each snippet and decides whether or not interrupting the backbone network and outputting the final result; (4) Discriminator outputs a final prediction result for a given snippet if the controlling agent decides to activate it. The overview of our proposed architecture is shown in Figure \ref{fig:framework}.

\subsection{Snippet Generator}
ECG signal is a special multivariate time series with periodic characteristics as it reflects physiology activities of the heart. To well analyze such kind of data, we propose the snippet generator, which splits the original time series into individual sub-series segments according to the periodicity. In detail for ECG data, the snippet can be generated as a heartbeat-based sub-time series, meaning that each snippet represents a single heartbeat. The input of the snippet generator is a time series sample $X^{(i)}$, and the output is the set of snippets $B^{(i)}$. $B^{(i)} = \{b_{1}^{i},b_{2}^{i},...,b_{T}^{i}\}$, where $b_{t}^{i}$ is a single snippet, $t$ means the index and $T$ is the number of snippet by $X^{(i)}$. By inheriting from $X^{(i)}$, each $b^{(i)} \in B^{(i)}$ shares the same label $y^{(i)}$. In this paper, our focus is not placed on heartbeat detection, but instead have adopted the existing method proposed in \mbox{\cite{christov2004real}}.  With reference to the QRS-detection method in \cite{christov2004real}, the heartbeats of each ECG signal are extracted and then sorted chronically.

\subsection{Backbone Network}

Backbone network aims to model the local spatial dependency of each snippet and the global temporal dependency across snippets to generate valuable hidden representations. A multi-layer convolutional network is adopted in the backbone network to learn the correlations between each variate of the input snippets and then capture the hidden spatial state. As the different scales of the ECG problem arise, the convolutional network's depth and design would be varied based on the specific target. It is worthy to notice that our proposed model was designed to be compatible with various convolutional neural networks. Moreover, to propagate the information across different snippets, a recurrent network, LSTM, is designed to leverage the hidden spatial state generated by the convolutional network and generate the spatial-temporal hidden state. 

\subsubsection{Convolutional Neural Networks Block}
The CNN block is designed to learn inner-snippet spatial dependency and output the snippet spatial hidden state $S_t$. Mainly, this block consists of convolution layers and pooling layers, which are responsible for extracting effectively hidden states from the snippet of ECG signals automatically. The convolution layer can be regarded as a fuzzy filter that captures the local spatial dependency of each snippet of the input ECG signals and reduces noises. The pooling layer aims to reduce the feature maps' dimensions and summarize critical features generated by a convolution layer. The specific network structure and parameters are detailed in Section 4. In short, the input snippet $b_{t}$ from an ECG signal $X$ is feed into the CNN block and then converted into a low-dimensional hidden state $S_{t}$. The output of the CNN block can be represented as follows:

\begin{equation}
    S_{t} = CNN(b_{t})
\end{equation}

\subsubsection{Long short-term Memory Block} 
Following the CNN block, a recurrent network extended with long-short term memory (LSTM)\mbox{\cite{hochreiter1997long}} cells propagates the snippet information, mapping spatial hidden states generated at each snippet to a hidden vector presentation that stands for temporal state information. The recurrent network learns to encode each hidden spatial state of a snippet to hidden temporal representations as a state vector. Hidden state vector $H_t$ is computed by combining the current snippet spatial hidden state $S_t$ and the previous hidden state $H_{(t-1)}$, hence the recurrent nature of the model. The LSTM block passes the information of $H_{(t-1)}$ and records the information of $S_t$. Therefore, $H_t$ is the spatial-temporal hidden vector state that comprehensively combines the inner-snippet spatial dependency and inter-snippet temporal dependency. The spatial-temporal hidden state $H_t$ is then passed on to the next module, called the controlling agent, to manipulate the output decision and the discriminator to output the prediction result. Specifically, the hidden state vector $H_t$ can be obtained as follows:

\begin{equation}
    H_{t} = LSTM (S_{t}, H_{t-1})
\end{equation}

\begin{flushleft}
where $S_t$ is the hidden spatial state produced by the convolutional neural networks block, $H_{t-1}$ denotes the cell’s output and state at time $t-1$ inside the LSTM cell, respectively.
\end{flushleft}

\subsection{Controlling Agent}
The controlling agent is the key component that controls the workflow, deciding whether the backbone network keeps on working or activates the discriminator to generate a prediction result. To achieve this goal, we adopt reinforcement learning techniques\mbox{\cite{10.5555/3009657.3009806}}, solving a Partially-Observable Markov Decision Process (POMDP)\mbox{\cite{Spaan2012}}. Each snippet's spatial-temporal hidden state of an input ECG signal can be seen as an individual state; each state's action is chosen by using the well-learned policy, and a reward is obtained according to the quality of the selected action. The controlling agent is trained by gradient-based policy learning in which the objective is to optimize long-term rewards according to the performance of the discriminator. Critical components for a given reinforcement learning framework are detailed below.

\subsubsection{State} 

In reinforcement learning, states are the representations of the task's current environment, which describes the current situation. In our case, the agent's state is the set of currently perceived snippets, $O_1,O_2,…,O_t$, essentially the outputs of the backbone network. Here, the spatial-temporal hidden state $H_t$ of $O_t$ that encodes spatial and temporal information is used as an observation. It is the core information that decides the selection of an action by the learned policy.

\subsubsection{Policy} 
The policy is the strategy that the controlling agent employs to determine the next action based on the current state. A good way to reduce the number of states is by using a neural network-based policy\mbox{\cite{1416876}}, where the inputs are states and outputs are actions. Here, the policy selects an action by the current state $H_t$,  $a_t = \pi_{\theta}( H_t ) $. As introduced in the previous section, $H_t$ is the low-dimensional hidden state encoded inner and inter-dependency for representing an ECG-signal snippet. In this work, we adopt a fully-connected neural network to approximate this policy function. The policy here is a function for mapping the current state $H_t$ to the parameterized distribution of the set of actions.

\subsubsection{Action} 
Action is what the controlling agent can do in each state. Actions control the whole process between the backbone network and discriminator as following situations arise: if $a_t=0$, the agent moves forward to the next observation, and keeps inputting the snippet into the backbone network, then acquires the spatial-temporal hidden state of the corresponding snippet. On the other hand, if $a_t=1$, the agent selects to interrupt the backbone network and then informs the discriminator to predict a label by feeding the hidden vector state $H_t$. Then the approximated probability $\pi_{t}$ fits a Bernoulli distribution, for sampling an action according to $P(p= 1) = \pi_{t}$. Once, based on the output from Bernoulli distribution, when the agent selects interrupting the backbone network or the observation series is finished $(t=T)$, $t$ is considered to be the interrupting time point $\tau$.

\subsubsection{Reward}
The reward is a measure to quantify the parameters of the current policy. To boost collaboration between the agent and the discriminator, it must observe the discriminator's returns and measure the discriminator's degree of success. Thus, when the discriminator gives a correct label, the current policy's reward is $r_t=1$; otherwise, the reward is called punishment, denoted as $r_t=-1$. In this paper, we consider long-term rewards in the proposed SPN. For example, if the controlling agent stops at a time point $(t=5)$, and it outputs a correct answer, the reward will be $5$. In this case, if the controlling agent produces an incorrect answer, the reward will be $-5$. Overall, the objective of this agent is to maximize the total reward.

\subsection{Discriminator}
The discriminator is the final module of our proposed framework, which aims to predict a label $\hat{y}$ of a given time series by mapping the hidden state $H_t$ into a lower-dimensional space using a fully-connected network. The resulting low dimensional vector is then normalized to label probabilities via the softmax function. The equation is shown below:

\begin{equation}
    \hat{y} = \mathop{\arg\max}_{i} P(Y=i | H_t)
\end{equation}

\section{Experimental Evaluation}
In this section, we use a public real-world 12-lead ECG dataset to conduct a series of experiments to evaluate the proposed model's performance.

\subsection{Experiment Setting}
\subsubsection{Environment}
The proposed model and the state-of-the-art methods are trained and tested on a server with a Xeon Gold CPU, 128GB memory, and a GPU card, Nvidia V100. This server runs an Ubuntu 18.04 system, and the models are implemented based on the PyTorch\cite{paszke2017automatic} 0.4.1.

\subsubsection{Data Source}
The ECG data used in this study was from the 1st China Physiological Signal Challenge \cite{liu2018open}. This dataset contains 6877 12-lead ECG records from 6 s to 60 s. A total of 8 types of arrhythmia (i.e., AF, I-AVB, LBBB, RBBB, PAC, PVC, STD, and STE) and normal sinus rhythms were to be classified in these records. These records were collected from 11 hospitals and sampled at 500 Hz. We run 10-fold cross-validation on this dataset for comparing the methods.

\subsubsection{Performance Metrics} 
In this research, typical classification metrics, including accuracy, earliness, precision, recall, and F1-score, are used for comparing our proposed model with the state-of-the-art methods. In addition, the harmonic mean is conducted to comprehensively evaluate the performance of the competing methods in balancing accuracy and earliness.

\textbf{Accuracy:} It is a performance evaluation measure that is the proportion of correct predictions and total predictions, formally defined as:

\begin{equation}
Accuracy = \frac{1}{m}\sum_{i=1}^{m}{(y_{i}==\hat{y}_{i})}
\end{equation}

\begin{flushleft}
where $m$ is the number of total sample instances in the testing set; $y_{i} $ and $ \hat{y}_{i} $ are the ground truth class and the predicted class for the $i^{th}$ sample, respectively. The accuracy of a model is calculated as the percentage of correct predictions of the test samples. Moreover,
\end{flushleft}

\begin{equation}
Earliness = \frac{1}{m}\sum_{i=1}^{m}{ \frac{s}{L}}
\end{equation}

\begin{flushleft}
where $s$ is the time point at prediction for a single time series made by a given eTSC model and $L$ is the time series's length.
\end{flushleft}

As noted in the introduction, eTSC thus has two naturally contradictory optimization goals. Hence, eTSC can be evaluated in different directions, comparing accuracies by keeping earliness constant or comparing earliness by keeping accuracy constant. To evaluate two goals simultaneously, a popular choice that appeared in the previous research \mbox{\cite{schafer2019teaser}} is the harmonic mean of earliness and accuracy. Harmonic mean (HM) is a metric for measuring the bi-objective optimization problems. In our case, we measure earliness and accuracy comprehensively as defined below.

\begin{equation}
    HM = \frac{ 2 \times (1-Earliness) \times Accuracy}{(1-Earliness)+Accuracy} 
\end{equation}

Moreover, the average precision, recall rate, and F1-score are adopted for measuring the multi-classification performances. The details are shown as follows:

\begin{equation}
    Precision = \frac{TP}{TP+FP}
\end{equation}

\begin{equation}
    Recall = \frac{TP}{TP+FN}
\end{equation}

\begin{equation}
    F1 = \frac{2 \times (Precision \times Recall) }{Precision + Recall}
\end{equation}

For a specific class in the multi-classification problems, TP (true positive) indicates the number of correctly classified samples, FN (false negative) is the number of samples that are misclassified into other classes, and FP (false positive) refers to the number of samples with the other class that misclassified in this class.

\subsubsection{Competing Methods}
We compare our model with the following state-of-the-art approaches. To experiment with these methods, we either use the authors' public codebase or follow their original paper to implement it to the best of our knowledge.

\begin{itemize}
    \item SR2-CF2 \cite{mori2017early}: This is a feature-based model for early time series classification, in which the features are generated according to a given distance function. Based on the gene algorithm, it outputs a confident classification time point.
    \item EARLIEST \cite{10.1145/3292500.3330974}:  This is a reinforcement learning-based approach. It outputs the classification result by a well-trained policy network.
    \item TEASER \cite{schafer2019teaser}: Based on a series of sub-classifiers for each time point, TEASER is proposed to deal with varied-length time series early classification problem.
    \item MDDNN \cite{huang2018multivariate}: This is a deep learning-based model that combines CNN and LSTM to solve the early time series classification problem.
    \item ETEeTSC \cite{russwurm2019end}: This is another deep learning-based early time series classification model. Based on a new loss function, this model can optimize accuracy and earliness simultaneously.
\end{itemize}

Note that we do not compare our methods with the classical works, such as ECTS \cite{xing2012early} and EDSC \cite{xing2011extracting}. The reason is that these methods appear to be out-of-date, and they were outperformed by the above models, SR2-CF2, TEASER, and ETEeTSC.

\subsubsection{Parameters Setting}
In the backbone convolutional neural networks of the proposed framework, 13 convolutional layers divided into five blocks are used for learning snippet spatial dependency. For each convolutional layer, a batch normalization layer (Batch-norm) \mbox{\cite{ioffe2015batch}} and a rectified linear unit (ReLU) function \mbox{\cite{nair2010rectified}} are adopted. The kernel size of 3, boundary padding of 1, and stride of 1 are set for all convolutional layers. Moreover, a pooling layer with kernel size 3 and stride size 3 controls the output size for each block. Therefore, the input length was maintained during convolutions and only adjusted by pooling layers for each block. A single LSTM layer with 256 cells is used for learning snippet temporal dependency in the long short-term memory block. In the training phase, we use the Adam optimizer \mbox{\cite{kingma2014adam}} training in each mini-batch and update the parameters. The learning rate is set as $10^{-3}$ and divided by five at every 20 epochs, eventually terminated at 100 epochs. All the training data is divided into mini-batches for network training, and the mini-batch size is set as 32.

\begin{table*}[!ht]

\caption{Performance Comparison on ICBEB dataset.}
\label{tab:icbeb}
\centering
\renewcommand{\arraystretch}{1.35}
\begin{tabular}{|c|c|c|c|c|c|c|} \hline \hline
& Accuracy & Earliness & Precision & Recall & F1-score & Harmonic Mean \\ 
\hline \hline

SR2-CF2 & 0.167 $\pm$ 0.009 & 0.228 $\pm$ 0.002 & 0.579 $\pm$ 0.110 & 0.160 $\pm$ 0.009 & 0.102 $\pm$ 0.011 & 0.274 $\pm$ 0.013 \\ \hline

EARLIEST & 0.283 $\pm$ 0.009 & \textbf{0.001 $\pm$ 0.001} & 0.176 $\pm$ 0.050 & 0.151 $\pm$ 0.011 & 0.115 $\pm$ 0.013 & 0.441 $\pm$ 0.010 \\ \hline

TEASER   & 0.456 $\pm$ 0.020 & 0.549 $\pm$ 0.009 & 0.409 $\pm$ 0.032 & 0.366 $\pm$ 0.015 & 0.368 $\pm$ 0.016 & 0.453 $\pm$ 0.012 \\ \hline

MDDNN    & 0.585 $\pm$ 0.015 & 0.455 $\pm$ 0.005 & 0.522 $\pm$ 0.016 & 0.511 $\pm$ 0.015 & 0.511 $\pm$ 0.015 & 0.564 $\pm$ 0.007 \\ \hline

ETEeTSC  & 0.735 $\pm$ 0.028 & 0.416 $\pm$ 0.037 & 0.704 $\pm$ 0.037 & 0.686 $\pm$ 0.033 & 0.687 $\pm$ 0.038 & 0.649 $\pm$ 0.022 \\ \hline

SPN & 0.796 $\pm$ 0.013 & 0.387 $\pm$ 0.011 & 0.769 $\pm$ 0.017 & 0.748 $\pm$ 0.016 & 0.751 $\pm$ 0.015 & 0.694 $\pm$ 0.008\\ 
\hline \hline

\end{tabular}%
\end{table*}

\subsection{Experiment Results}
\subsubsection{Performance Comparison} \label{sec:performance}
Table \mbox{\ref{tab:icbeb}} shows the CVDs classification performance of the proposed method and competing models for the 12-lead ECG dataset.  As shown in Table \mbox{\ref{tab:icbeb}}, our proposed model achieves the CVDs classification task's best performance because it can simultaneously learn the inner-spatial and inter-temporal dependencies. Considering the feature-based models, SR2-CF2 and TEASER, the results demonstrate that these approaches cannot solve ECG early classification satisfactorily. The reason is that these methods are designed and evaluated in the short length time-series datasets, in which they cannot acquire a good result while handling long time series. On the other hand, deep-learning-based methods, such as MDDNN and ETEeTSC, show acceptable accuracy for the classification task, but these methods' earliness metric does not look good. EARLIEST is the first reinforcement learning-based model for early time series classification. It shows good earliness performance, but the poor classification performance largely due to its premature decision because of its overly simplified architecture. Overall, our proposed method presents the best result. It is worthy to note that around $7\%$ classification performance improvement is gained by our proposed method compared to the state-of-the-art methods in terms of precision, recall, accuracy, and F1-score.

As discussed in the related work section, since SR2-CF2 is a feature-based model, the features for its classifier are extracted based on a given distance function. In general, it cannot work well on the long and varied length time series classification. The same problem also appears in the experiments of the EARLIEST and TEASER. These models solve the trade-off between accuracy and earliness in the early time classification problem, but the design of these approaches does not consider the long-length time series data. For processing long time series, the weight parameters between accuracy and earliness of these models are inappropriately set due to the penalty of time being much larger than the penalty of accuracy. For MDDNN and ETEeTSC, their performances are better than others as they exploit deep neural architecture, while ETEeTSC has a unique ability to output the prediction time point automatically. In contrast, SPN solves the ECG early classification problem with long-length and varied-length time-series properties well, and the result is remarkable, meeting all real-world application criteria.

\section{Conclusion}
This paper has addressed the problem of cardiovascular disease early classification based on varied-length multi-lead ECG. It is a critical real-world application, and this problem has not been well studied in the research community. For solving this problem, a novel deep reinforcement learning framework, Snippet Policy Network, was proposed consisting of four modules, snippet generator, backbone network, controlling agent, and discriminator. The backbone network is proposed to learn the inner-snippet spatial correlations and inter-snippet temporal correlations by combining convolutional and recurrent network architectures. A controlling agent with reinforcement architecture is proposed to solve the earliness and accuracy conflicting goals of bi-objective optimization. This agent allows the model to classify the CVDs accurately and search for the earlier prediction time point. The discriminator is proposed to make a classification result by mapping the features generated by the backbone network. Through a series of experiments, the results demonstrate that our proposed model outperforms the state-of-the-art methods at least $7\%$ in terms of precision, recall, accuracy, F1-score, and harmonic mean. Moreover, our model achieves high accuracy result of more than $80\%$ for multiple disease classifications, filling the gap between the research community and medical practices. All in all, Snippet Policy Network presents a valid solution to the problem of cardiovascular disease early classification based on varied-length multi-lead ECG.

In our future work, we aim to improve the agent mechanism and explore the Snippet Policy Network's data interpretability. We believe that data interpretability may help the medical professional better understand the principle of CVDs classification.

\ifCLASSOPTIONcaptionsoff
  \newpage
\fi

\bibliographystyle{IEEEtran}
\bibliography{references}

\begin{thebibliography}{10}
\providecommand{\url}[1]{#1}
\csname url@samestyle\endcsname
\providecommand{\newblock}{\relax}
\providecommand{\bibinfo}[2]{#2}
\providecommand{\BIBentrySTDinterwordspacing}{\spaceskip=0pt\relax}
\providecommand{\BIBentryALTinterwordstretchfactor}{4}
\providecommand{\BIBentryALTinterwordspacing}{\spaceskip=\fontdimen2\font plus
\BIBentryALTinterwordstretchfactor\fontdimen3\font minus
  \fontdimen4\font\relax}
\providecommand{\BIBforeignlanguage}[2]{{%
\expandafter\ifx\csname l@#1\endcsname\relax
\typeout{** WARNING: IEEEtran.bst: No hyphenation pattern has been}%
\typeout{** loaded for the language `#1'. Using the pattern for}%
\typeout{** the default language instead.}%
\else
\language=\csname l@#1\endcsname
\fi
#2}}
\providecommand{\BIBdecl}{\relax}
\BIBdecl

\bibitem{van2004clinical}
C.~Van~Mieghem, M.~Sabbe, and D.~Knockaert, ``The clinical value of the ecg in
  noncardiac conditions,'' \emph{Chest}, vol. 125, no.~4, pp. 1561--1576, 2004.

\bibitem{yao2020multi}
Q.~Yao, R.~Wang, X.~Fan, J.~Liu, and Y.~Li, ``Multi-class arrhythmia detection
  from 12-lead varied-length ecg using attention-based time-incremental
  convolutional neural network,'' \emph{Information Fusion}, vol.~53, pp.
  174--182, 2020.

\bibitem{alkukhun2014electrocardiographic}
L.~Alkukhun, M.~Baumgartner, M.~Budev, R.~A. Dweik, and A.~R. Tonelli,
  ``Electrocardiographic differences between copd patients evaluated for lung
  transplantation with and without pulmonary hypertension,'' \emph{COPD},
  vol.~11, no.~6, pp. 670--680, 2014.

\bibitem{GOUDIS2015264}
C.~A. Goudis, A.~K. Konstantinidis, I.~V. Ntalas, and P.~Korantzopoulos,
  ``Electrocardiographic abnormalities and cardiac arrhythmias in chronic
  obstructive pulmonary disease,'' \emph{Int. J. Cardiol.}, vol. 199, pp. 264
  -- 273, 2015.

\bibitem{wang2016earliness}
\BIBentryALTinterwordspacing
W.~Wang, C.~Chen, W.~Wang, P.~Rai, and L.~Carin, ``Earliness-aware deep
  convolutional networks for early time series classification,'' \emph{CoRR},
  vol. abs/1611.04578, 2016. [Online]. Available:
  \url{http://arxiv.org/abs/1611.04578}
\BIBentrySTDinterwordspacing

\bibitem{mori2017reliable}
U.~Mori, A.~Mendiburu, E.~Keogh, and J.~A. Lozano, ``Reliable early
  classification of time series based on discriminating the classes over
  time,'' \emph{Data Min Knowl Discov}, vol.~31, no.~1, pp. 233--263, 2017.

\bibitem{xing2012early}
Z.~Xing, J.~Pei, and S.~Y. Philip, ``Early classification on time series,''
  \emph{Knowl Inf Syst}, vol.~31, no.~1, pp. 105--127, 2012.

\bibitem{xing2011extracting}
Z.~Xing, J.~Pei, P.~S. Yu, and K.~Wang, ``Extracting interpretable features for
  early classification on time series,'' in \emph{Proceedings of the 2011 SIAM
  International Conference on Data Mining}.\hskip 1em plus 0.5em minus
  0.4em\relax SIAM, 2011, pp. 247--258.

\bibitem{ghalwash2012early}
M.~F. Ghalwash and Z.~Obradovic, ``Early classification of multivariate
  temporal observations by extraction of interpretable shapelets,'' \emph{BMC
  Bioinform.}, vol.~13, no.~1, p. 195, 2012.

\bibitem{ghalwash2013extraction}
M.~F. Ghalwash, V.~Radosavljevic, and Z.~Obradovic, ``Extraction of
  interpretable multivariate patterns for early diagnostics,'' in
  \emph{Proceedings of the IEEE 13th International Conference on Data Mining},
  ser. ICDM '13.\hskip 1em plus 0.5em minus 0.4em\relax IEEE, 2013, pp.
  201--210.

\bibitem{he2019ensemble}
G.~He, W.~Zhao, X.~Xia, R.~Peng, and X.~Wu, ``An ensemble of shapelet-based
  classifiers on inter-class and intra-class imbalanced multivariate time
  series at the early stage,'' \emph{Soft Comput.}, vol.~23, no.~15, pp.
  6097--6114, 2019.

\bibitem{parrish2013classifying}
N.~Parrish, H.~S. Anderson, M.~R. Gupta, and D.~Y. Hsiao, ``Classifying with
  confidence from incomplete information,'' \emph{J Mach Learn Res}, vol.~14,
  no.~1, p. 3561–3589, Dec. 2013.

\bibitem{mori2017early}
U.~Mori, A.~Mendiburu, S.~Dasgupta, and J.~A. Lozano, ``Early classification of
  time series by simultaneously optimizing the accuracy and earliness,''
  \emph{IEEE Trans Neural Netw Learn Syst}, vol.~29, no.~10, pp. 4569--4578,
  2017.

\bibitem{schafer2019teaser}
P.~Sch{\"a}fer and U.~Leser, ``Teaser: Early and accurate time series
  classification,'' p. 1336–1362, 2020.

\bibitem{huang2018multivariate}
H.-S. Huang, C.-L. Liu, and V.~S. Tseng, ``Multivariate time series early
  classification using multi-domain deep neural network,'' in \emph{Proceedings
  of the IEEE 5th International Conference on Data Science and Advanced
  Analytics}, ser. DSAA '18.\hskip 1em plus 0.5em minus 0.4em\relax IEEE, 2018,
  pp. 90--98.

\bibitem{martinez2018deep}
C.~Martinez, G.~Perrin, E.~Ramasso, and M.~Rombaut, ``A deep reinforcement
  learning approach for early classification of time sseries,'' in
  \emph{Proceedings of the 26th European Signal Processing Conference}, ser.
  EUSIPCO '18.\hskip 1em plus 0.5em minus 0.4em\relax IEEE, 2018, pp.
  2030--2034.

\bibitem{hsu2019multivariate}
E.-Y. Hsu, C.-L. Liu, and V.~S. Tseng, ``Multivariate time series early
  classification with interpretability using deep learning and attention
  mechanism,'' in \emph{Proceedings of the 23th Pacific-Asia Conference on
  Knowledge Discovery and Data Mining}.\hskip 1em plus 0.5em minus 0.4em\relax
  Springer, 2019, pp. 541--553.

\bibitem{10.1145/3292500.3330974}
T.~Hartvigsen, C.~Sen, X.~Kong, and E.~Rundensteiner, ``Adaptive-halting policy
  network for early classification,'' in \emph{Proceedings of the 25th ACM
  International Conference on Knowledge Discovery and Data Mining}, ser. KDD
  ’19.\hskip 1em plus 0.5em minus 0.4em\relax New York, NY, USA: Association
  for Computing Machinery, 2019, p. 101–110.

\bibitem{russwurm2019end}
\BIBentryALTinterwordspacing
M.~Ru{\ss}wurm, S.~Lef{\`{e}}vre, N.~Courty, R.~Emonet, M.~K{\"{o}}rner, and
  R.~Tavenard, ``End-to-end learning for early classification of time series,''
  \emph{CoRR}, vol. abs/1901.10681, 2019. [Online]. Available:
  \url{http://arxiv.org/abs/1901.10681}
\BIBentrySTDinterwordspacing

\bibitem{christov2004real}
I.~I. Christov, ``Real time electrocardiogram qrs detection using combined
  adaptive threshold,'' \emph{Biomed. Eng. Online}, vol.~3, no.~1, pp. 1--9,
  2004.

\bibitem{hochreiter1997long}
S.~Hochreiter and J.~Schmidhuber, ``Long short-term memory,'' \emph{Neural
  Comput.}, vol.~9, no.~8, p. 1735–1780, Nov. 1997.

\bibitem{10.5555/3009657.3009806}
R.~S. Sutton, D.~McAllester, S.~Singh, and Y.~Mansour, ``Policy gradient
  methods for reinforcement learning with function approximation,'' in
  \emph{Proceedings of the 12th International Conference on Neural Information
  Processing Systems}, ser. NIPS’99.\hskip 1em plus 0.5em minus 0.4em\relax
  Cambridge, MA, USA: MIT Press, 1999, p. 1057–1063.

\bibitem{Spaan2012}
M.~T.~J. Spaan, \emph{Partially observable markov secision processes}.\hskip
  1em plus 0.5em minus 0.4em\relax Berlin, Heidelberg: Springer Berlin
  Heidelberg, 2012, pp. 387--414.

\bibitem{1416876}
G.~G. {Yen} and P.~G. {DeLima}, ``Improving the performance of globalized dual
  heuristic programming for fault tolerant control through an online learning
  supervisor,'' \emph{IEEE Trans. Autom. Sci. Eng.}, vol.~2, no.~2, pp.
  121--131, 2005.

\bibitem{paszke2017automatic}
A.~Paszke, S.~Gross, F.~Massa, A.~Lerer, J.~Bradbury, G.~Chanan, T.~Killeen,
  Z.~Lin, N.~Gimelshein, L.~Antiga, A.~Desmaison, A.~Kopf, E.~Yang, Z.~DeVito,
  M.~Raison, A.~Tejani, S.~Chilamkurthy, B.~Steiner, L.~Fang, J.~Bai, and
  S.~Chintala, ``Pytorch: An imperative style, high-performance deep learning
  library,'' in \emph{Proceedings of the 33rd International Conference on
  Neural Information Processing Systems}.\hskip 1em plus 0.5em minus
  0.4em\relax Curran Associates, Inc., 2019, pp. 8024--8035.

\bibitem{liu2018open}
F.~Liu, C.~Liu, L.~Zhao, X.~Zhang, X.~Wu, X.~Xu, Y.~Liu, C.~Ma, S.~Wei, Z.~He
  \emph{et~al.}, ``An open access database for evaluating the algorithms of
  electrocardiogram rhythm and morphology abnormality detection,'' \emph{J Med
  Imaging Health Inform}, vol.~8, no.~7, pp. 1368--1373, 2018.

\bibitem{ioffe2015batch}
S.~Ioffe and C.~Szegedy, ``Batch normalization: Accelerating deep network
  training by reducing internal covariate shift,'' p. 448–456, 2015.

\bibitem{nair2010rectified}
V.~Nair and G.~E. Hinton, ``Rectified linear units improve restricted boltzmann
  machines,'' in \emph{Proceedings of the 27th International Conference on
  International Conference on Machine Learning}, ser. ICML’10.\hskip 1em plus
  0.5em minus 0.4em\relax Madison, WI, USA: Omnipress, 2010, p. 807–814.

\bibitem{kingma2014adam}
D.~P. Kingma and J.~Ba, ``Adam: A method for stochastic optimization,''
  \emph{arXiv preprint arXiv:1412.6980}, 2014.

\end{thebibliography}

\end{document}